\documentclass[10pt,twocolumn,letterpaper]{article}

\usepackage[pagenumbers]{cvpr}

\definecolor{cvprblue}{rgb}{0.21,0.49,0.74}
\usepackage[pagebackref,breaklinks,colorlinks,allcolors=cvprblue]{hyperref}
\usepackage{algorithm}
\usepackage[noend]{algpseudocode}
\algrenewcommand\algorithmicrequire{\textbf{Input:}}
\algrenewcommand\algorithmicensure{\textbf{Output:}}

\usepackage{booktabs}
\usepackage{tabularx}
\usepackage{multirow}
\usepackage{makecell}
\usepackage{threeparttable}
\usepackage{xcolor,colortbl}
\usepackage{multirow}
\usepackage{lipsum}
\usepackage{placeins}

\usepackage{algorithm}
\usepackage{algpseudocode}
\usepackage{float} 
\algrenewcommand\algorithmicrequire{\textbf{Input:}}
\algrenewcommand\algorithmicensure{\textbf{Output:}}
\usepackage{xcolor}

\definecolor{topcDark}{RGB}{102,113,246}   
\definecolor{topcLight}{RGB}{230,220,252}  

\title{Deep Forcing: Training-Free Long Video Generation \\
with Deep Sink and Participative Compression}

\author{Jung Yi\quad
Wooseok Jang\quad
Paul Hyunbin Cho\quad 
Jisu Nam\quad
Heeji Yoon\quad 
Seungryong Kim\\[10pt]
KAIST AI\\[3pt]
{\tt \href{https://cvlab-kaist.github.io/DeepForcing/}{\textcolor{purple}{https://cvlab-kaist.github.io/DeepForcing}}}
}

\begin{document}

\twocolumn[{%
 \renewcommand\twocolumn[1][]{#1}%
 \maketitle
 \centering
 \includegraphics[width=\textwidth, trim={0 0.5cm 0 0.5cm}]{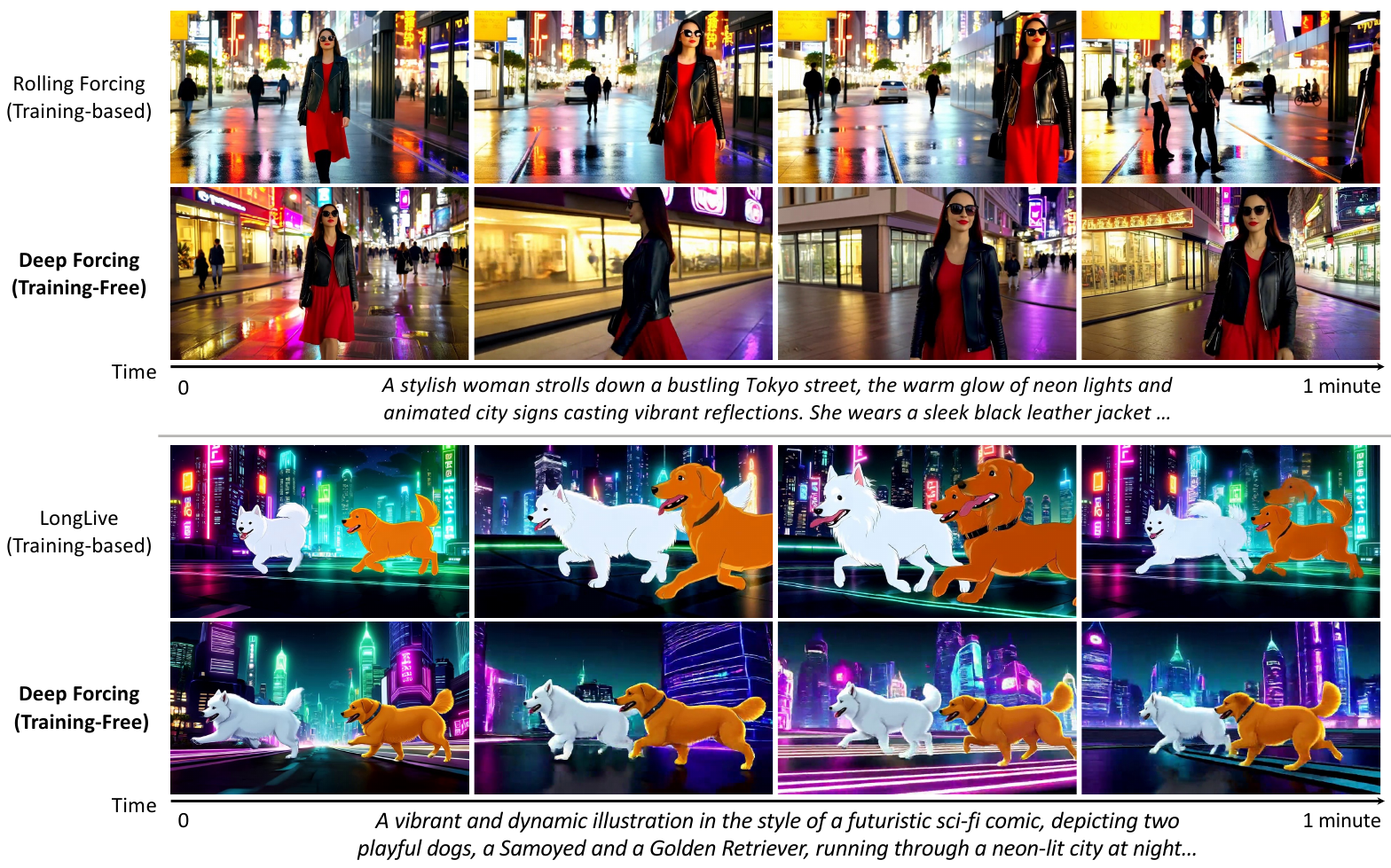}
 \vspace{-10pt}
 \captionof{figure}{Our \textbf{\textit{training-free}} approach, \textbf{Deep Forcing}, achieves comparable visual quality to \textbf{\textit{training-based}} baselines, such as Rolling Forcing~\cite{liu2025rolling} and LongLive~\cite{yang2025longlive}. Notably, Deep Forcing enables minute-long video generation while maintaining visual quality and dynamics \textbf{\textit{without}} requiring any additional training.
    \label{fig:teaser}
   }\vspace{+10pt}
}]

\maketitle
\begin{abstract}
Recent advances in autoregressive video diffusion have enabled real-time frame streaming, yet existing solutions still suffer from temporal repetition, drift, and motion deceleration. 
We find that naïvely applying StreamingLLM-style attention sinks to video diffusion leads to fidelity degradation and motion stagnation.
To overcome this, we introduce \textbf{Deep Forcing}, which consists of two \textbf{training-free} mechanisms that address this without any fine-tuning. Specifically, 1) \textbf{Deep Sink} dedicates half of the sliding window to persistent sink tokens and re-aligns their temporal RoPE phase to the current timeline, stabilizing global context during long rollouts. 2) \textbf{Participative Compression} performs importance-aware KV cache pruning that preserves only tokens actively participating in recent attention while safely discarding redundant and degraded history, minimizing error accumulation under out-of-distribution length generation. Together, these components enable \textbf{over 12$\times$ extrapolation} (e.g. \textbf{5s-trained $\rightarrow$ 60s+} generation) with better imaging quality than LongLive, better aesthetic quality than RollingForcing, almost maintaining overall consistency, and substantial gains in dynamic degree, all while maintaining real-time generation. Our results demonstrate that training-free KV-cache management can match or exceed training-based approaches for autoregressively streaming long-video generation. 
\end{abstract}

\section{Introduction}
\label{sec:intro}
Recent advances in video diffusion models~\cite{yang2024cogvideox, wan2025, HaCohen2024LTXVideo} have demonstrated remarkable capabilities in synthesizing short video clips (e.g, 50--81 frames) with both high visual fidelity and coherent motion dynamics. 

Building upon this progress, emerging interactive systems, such as world models~\cite{rre, rre2, yan}, now require autoregressive video generation (e.g., 1-2 minutes), where frames are streamed sequentially in real time for immediate downstream applications~\cite{shin2025motionstream, yan, rre}. Unlike conventional offline video generation, which synthesizes entire clips at once, autoregressive video generation operates in an online manner, with each frame generated, emitted, and consumed instantaneously. Self Forcing~\cite{huang2025self} and its variants~\cite{yang2025longlive,liu2025rolling,cui2025self} have become standard in this field, leveraging a causal attention mask and the Key-Value (KV) cache from previous frames.

However, this existing autoregressive formulation is inherently susceptible to \textit{error accumulation over long horizons}, as each predicted frame depends on previously generated and potentially imperfect frames~\cite{yin2025slow, huang2025self, wang2025error}. Such accumulation error leads to fidelity degradation, in which visual quality deteriorates as colors drift toward over-saturation, textures blur, and fine details disappear.

To mitigate this, several works~\cite{chen2024diffusion, song2025history} introduce history corruption by adding noise to previous frames during training. While this improves robustness to noisy generated histories at inference, a discrepancy remains between generated noise and artificially injected noise, leaving models vulnerable to long-horizon drift. Self Forcing~\cite{huang2025self} aims to reduce this gap by training on self-generated histories, but their heavy reliance on past frames’ KV caches still leads to accumulated errors. 

On the other hand, recent LLM studies~\cite{Ghadia2025DialogueWL, shilacache, li2024snapkv}, inspired by StreamingLLM~\cite{xiao2023efficient}, introduce an \textit{attention sink}, in which newly generated tokens attend strongly to a small set of initial global tokens, helping stabilize the attention distribution and improve overall performance.

Motivated by these insights, we propose \textbf{Deep Forcing}, a novel \textbf{tuning-free} method that addresses error accumulation in long-horizon video generation. Our approach is able to generate minute-long video that maintain both visual fidelity and motion stability, while requiring no fine-tuning, outperforming even training-based approaches~\cite{liu2025rolling, yang2025longlive} (Fig.~\ref{fig:teaser}).
\begin{figure}[t]
  \centering
   \includegraphics[width=1.0\linewidth]{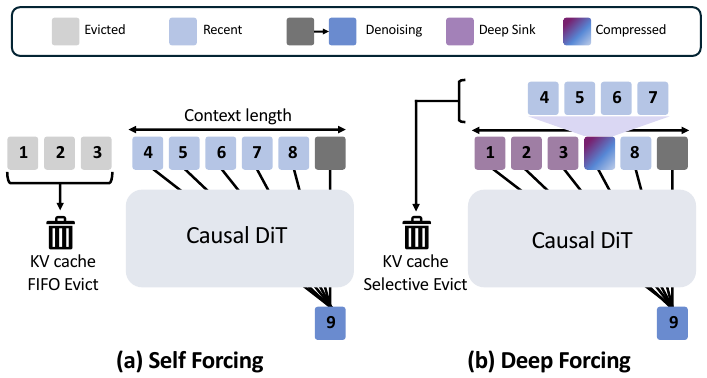}
   \vspace{-20pt}
   \caption{\textbf{Comparison of KV Cache Management.}
(a) \textbf{Self Forcing}~\cite{huang2025self} adopts a FIFO policy that discards the earliest tokens regardless of their importance, often losing critical context and degrading generation quality.
In contrast, our (b) \textbf{Deep Forcing} performs selective eviction by preserving \textbf{Deep Sink} tokens and applying KV-cache compression, effectively mitigating visual degradation during long-horizon generation.}
   \label{fig:abstract_comp}\vspace{-10pt}
\end{figure}

Specifically, we observe that the pre-trained Self Forcing~\cite{huang2025self} inherently exhibits strong attention sink behavior, not only attending to the first few tokens but also strongly attending to intermediate tokens. Building on this finding, we first introduce \textbf{Deep Sink}, which (i) maintains a large deep-sink ratio (typically $40$--$60\%$) and (ii) dynamically adjusts the Relative Positional Embedding (RoPE) for long video generation. Second, we further present \textbf{Participative Compression} that retains only the most informative tokens in the KV cache by selecting them based on their importance to queries from recent frames. This removes redundancy and significantly reduces fidelity degradation caused by noise accumulation from outdated tokens.

We implement our method on top of the pre-trained Self Forcing. Comprehensive evaluations using VBench~\cite{huang2024vbench}, user studies, and VLM assessments demonstrate that our training-free approach significantly enhances the baseline without any fine-tuning. We also achieve state-of-the-art performance on several metrics, even surpassing training-based methods~\cite{liu2025rolling, yang2025longlive}. Our ablation studies further validate the effectiveness of each design choice.

Our contribution is summarized as follows:
\begin{itemize}
\item We propose a \textbf{tuning-free} autoregressive video generation framework, dubbed \textbf{Deep Forcing}, that significantly mitigates error accumulation in long-horizon generation.
\item We introduce \textbf{Deep Sink}, which stabilizes long-horizon generation by leveraging the inherent attention sink behavior of Self Forcing~\cite{huang2025self} while adjusting the relative positional gap.
\item We present \textbf{Participative Compression}, a lightweight KV selection mechanism that removes redundant tokens.
\item Our training-free method achieves state-of-the-art performance on VBench and user studies, surpassing existing training-based approaches.
\end{itemize}

\begin{figure*}
    \centering
    \includegraphics[width=1.0\linewidth]{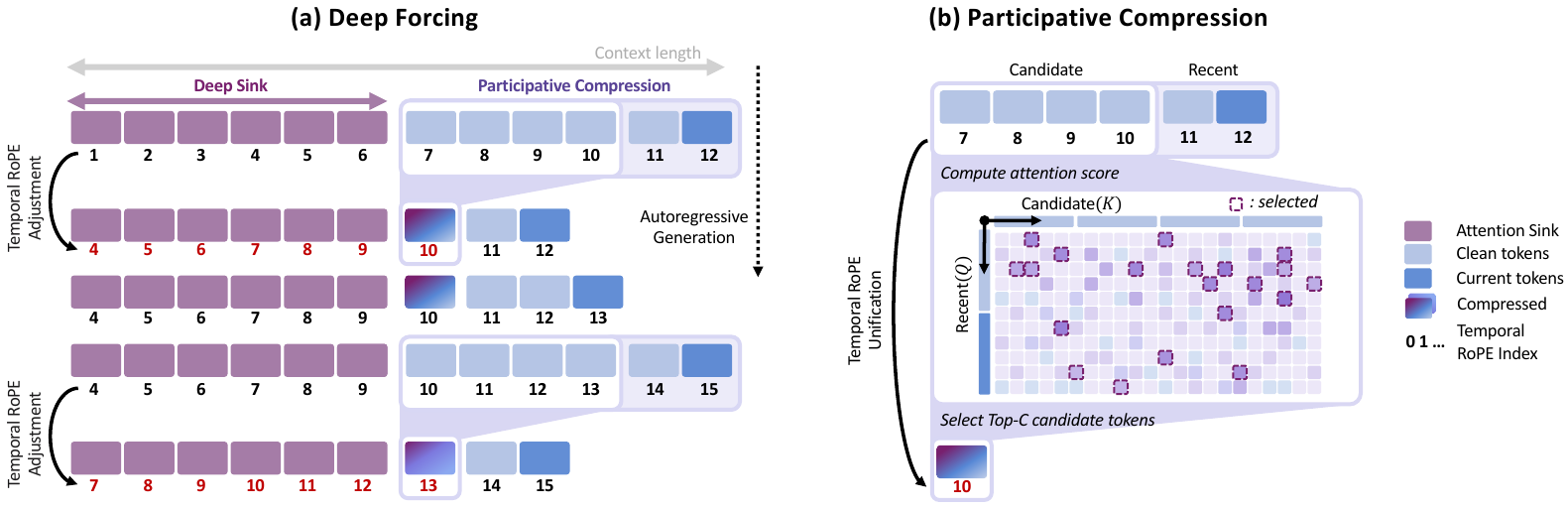} 
    \caption{
    \textbf{Overview of Deep Forcing.} \textbf{(a) Deep Forcing} maintains a substantially enlarged attention sink (Deep Sink) covering approximately half the context window, combined with Participative Compression for the remaining rolling portion. Temporal RoPE adjustment aligns the sink tokens' temporal indices with current frames to maintain temporal coherence. \textbf{(b) Participative Compression} computes query-averaged attention scores between recent tokens and candidate tokens, selecting the top-C most important tokens to retain in the compressed cache while evicting redundant tokens.
    }
    \label{fig:architecture}\vspace{-10pt}
\end{figure*}

\section{Related Work}
\label{sec:formatting}
\paragraph{Autoregressive Video Diffusion.}
A growing line of work~\cite{jin2024pyramidal,chen2024diffusion,teng2025magi,huang2025self,yin2025slow} combines diffusion modeling with autoregressive (AR) prediction to support long-horizon or streaming video generation. MAGI-1~\cite{teng2025magi} generates videos chunk-by-chunk autoregressively with progressive denoising, enabling streaming generation. CausVid~\cite{yin2025slow} converts a pre-trained bidirectional diffusion transformer into a causal AR generator with KV caching. Building on these ideas, Self Forcing~\cite{huang2025self} addresses the train–inference mismatch by conditioning the model on its own generated frames. Rolling Forcing~\cite{liu2025rolling} proposes expanding the diffusion window, and LongLive~\cite{yang2025longlive} incorporates KV recaching to maintain visual continuity while ensuring prompt adherence across scene transitions. In contrast, our method is fully tuning-free. We show that the pre-trained Self Forcing already has attention-sink behavior and demonstrate how to leverage it effectively to surpass existing training-based methods.

\paragraph{Attention Sink.}
Recent work has revealed that attention in autoregressive models concentrates disproportionately on initial tokens, termed attention sinks~\cite{xiao2023efficient}. 
StreamingLLM~\cite{xiao2023efficient} showed that retaining these sink tokens within a sliding window enables robust generation beyond the training context length.

Building on this insight, recent autoregressive video models~\cite{yang2025longlive,liu2025rolling} maintain the first three frames as attention sinks via model distillation or fine-tuning.
We demonstrate that the pre-trained autoregressive video diffusion model~\cite{huang2025self} exhibits inherent attention sink behavior that can be effectively leveraged without training, requiring deeper context preservation.

\paragraph{KV Cache Compression.}
The linearly growing KV cache in autoregressive generation motivates compression strategies that reduce memory footprint while preserving generation quality. As the cache grows, attention becomes distributed across increasingly many tokens, diluting focus on critical context and degrading output quality.
To address this, recent works employ attention-based token selection for long-context LLM generation. H2O~\cite{zhang2023h2o} and SnapKV~\cite{li2024snapkv} preserve important tokens based on cumulative attention scores and observation windows, respectively. D2O~\cite{wan2024d2o} dynamically allocates budgets across layers, while MorphKV~\cite{Ghadia2025DialogueWL} maintains constant-size caches through correlation-aware ranking. While these methods target language models, similar memory constraints arise in autoregressive video diffusion, where temporal context must be efficiently maintained across frames. We extend these principles through Participative Compression.

\section{Preliminaries}
\paragraph{Autoregressive Video Diffusion.}
Autoregressive video diffusion models~\cite{teng2025magi, chen2025skyreelsv2infinitelengthfilmgenerative, huang2025self} produce each frame or chunk conditioned on previously generated frames within a denoising diffusion process.

Given a video sequence of $N$ frames $x^{1:N} = (x^1, x^2, \dots, x^N)$, the autoregressive model applies the chain rule to factorize the joint distribution as
\begin{equation}
p(x^{1:N}) = \prod_{i=1}^{N} p(x^i \mid x^{<i}).
\end{equation}
A diffusion model parameterizes each conditional $p(x^i \mid x^{<i})$, generating the $i$-th frame by conditioning on previously generated frames $x^{<i} = (x^1, x^2, \dots, x^{i-1})$.

\paragraph{Self Forcing.}
Self Forcing~\cite{huang2025self} generates videos in an autoregressive manner using a rolling KV cache mechanism, producing frames or frame chunks sequentially, enabling efficient long video generation.
Each frame is encoded into latent tokens through the VAE encoder.
The method maintains a fixed-size cache of length $L$ that stores key-value pairs corresponding to the most recent $L$ frames. When the cache reaches capacity, the oldest entry is evicted to accommodate new frames, thereby maintaining a sliding context window over the $L$ most recent frames. During generation, self-attention is computed between queries from the frame(s)  currently being generated and the keys and values of cached context frames.
Specifically, Self Forcing employs a 4-step denoising process with noise schedule $\{t_0=0, t_1=250, t_2 =500, t_3=750, t_{4}=1{,}000\}$ ($T=4$), totaling 5 noise levels. Each frame $i$ is denoised iteratively across these timesteps. At denoising step $j$, the model processes a noisy intermediate frame $x^i_{t_j}$, conditioned on the KV cache of previously generated clean frames. The predicted clean frame is then perturbed with Gaussian noise at a lower noise level $t_{j-1}$ via the forward diffusion process $\Psi$, producing $x^i_{t_{j-1}}$ for the next denoising iteration. Formally, this process is expressed as:
\begin{equation}
x^i_{t_{j-1}} = \Psi\big(G_\theta(x^i_{t_j}, t_j, KV), t_{j-1}\big),
\end{equation}
where $x^i_{t_4} \sim \mathcal{N}(0, I)$ and $KV$ denotes the KV cache from previously generated frames.

\begin{figure}[t]
  \centering
   \includegraphics[width=0.9\linewidth]{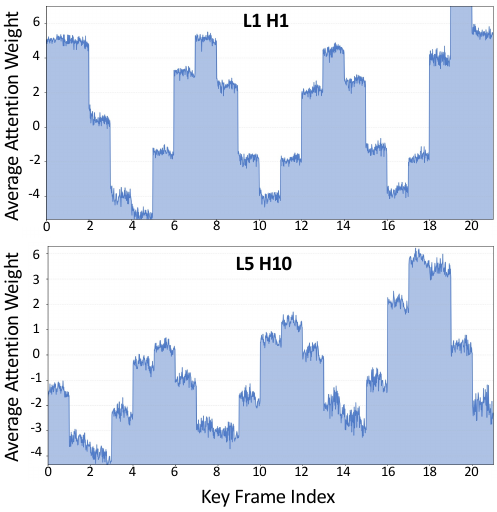}

   \caption{\textbf{Attention weight distribution across earlier frames.} Query-averaged attention showing how the last chunk (frames 19-21) attends to earlier KV cache entries (frames 0-18). We visualize two representative attention heads from different layers—L1H1 (layer 1, head 1) and L5H10 (layer 5, head 10)—demonstrating that substantial attention is maintained across the entire context window, not just initial frames. See Appendix~\ref{sec:additional_attn_vis} for additional heads analysis.}
   \label{fig:attention_analysis}
\end{figure}

\section{Method}
\subsection{Overview}
\label{sec:overview}
We propose a novel training-free method to mitigate error accumulation in long-horizon video generation.  
Drawing inspiration from the attention sink mechanism in large language models (LLMs)~\cite{xiao2023efficient}, our work begins by thoroughly investigating the attention mechanism within the pre-trained Self Forcing~\cite{huang2025self}. Based on this investigation, we introduce two core components: \textbf{Deep Sink}, which maintains approximately half of the sliding window as attention sinks, and \textbf{Participative Compression}, which selectively retains important tokens in the KV cache, while evicting redundant ones. Our method is illustrated in Fig.~\ref{fig:architecture}

\subsection{Deep Sink}
\label{sec:deep-sink}

\paragraph{Motivation.} 
Self Forcing \cite{huang2025self} employs a sliding window to autoregressively extrapolate video frames. However, because the model is distilled from short video clips (e.g., 5-second segments), its frame fidelity deteriorates significantly when generating sequences that extend far beyond its training domain.

This degradation is a known challenge in autoregressive systems. In the LLM domain, the attention sink mechanism~\cite{xiao2023efficient} was introduced as a simple yet effective technique to mitigate performance drift during sliding-window inference. 
While several works \cite{liu2025rolling, yang2025longlive} have investigated adapting the attention sink mechanism to redistribute attention probabilities and stabilize LLM performance, no prior work has explored how to achieve a similar stabilizing effect in autoregressive video diffusion models in a ~\textbf{training-free} manner.

To address this gap, we first analyze the attention behavior of the pre-trained Self Forcing. As illustrated in Fig.~\ref{fig:attention_analysis}, we specifically examine how newly generated latent frames attend to earlier frames in the KV cache. Contrary to the conventional understanding~\cite{xiao2023efficient} that only a small set of initial KV tokens (latent frames) needs to be retained, our analysis reveals: most attention heads allocate substantial weight to not only the earliest tokens, but also assign comparable attention to the middle of the sequence. 

\begin{figure}[t]
  \centering
   \includegraphics[width=1.0\linewidth]{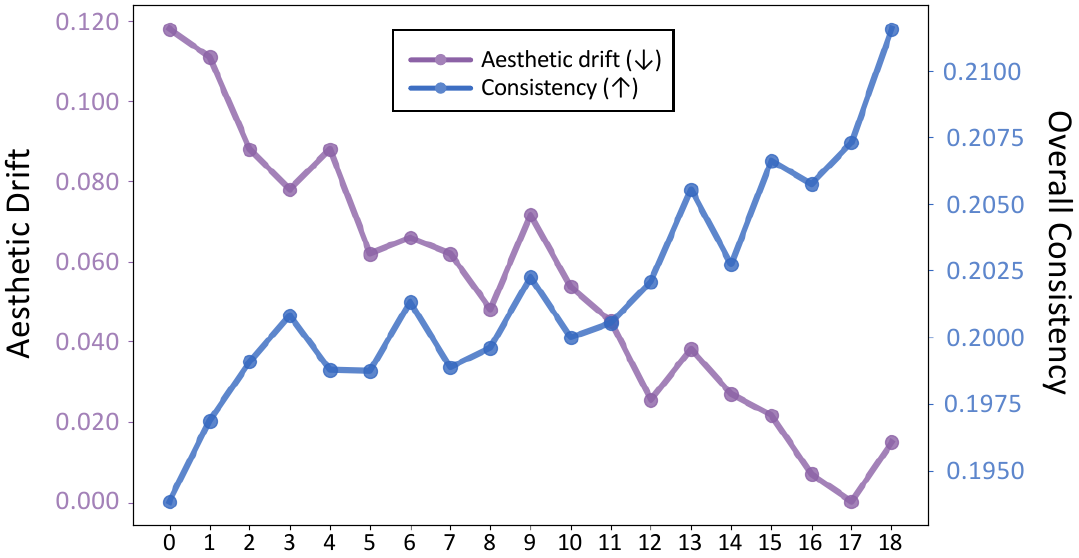}

   \caption{\textbf{Ablation study on Deep Sink depth.} We evaluate the effect of sink depth on video quality using Aesthetic Drift ($\downarrow$) and Overall Consistency ($\uparrow$) metrics on 50-second videos from the first 21 prompts in MovieGen~\cite{polyak2024movie}.}
   \label{fig:sink_depth_abl}
\end{figure}

\paragraph{Deepening Sink Size.} 
Based on this observation, we hypothesize that more tokens up to the middle of the initial sequence are essential for high-quality video generation. To evaluate this hypothesis, we measured the influence of different attention sink sizes on the generation quality of long videos. To rigorously assess long-horizon generation quality, we first define our key metrics from VBench~\cite{huang2024vbench}: Overall Consistency and Aesthetic Quality, which use ViCLIP~\cite{wang2023internvid} and LAION aesthetic predictor~\cite{laion}, respectively. Following standard practice in long video generation~\cite{zhang2025packing, yin2025slow, liu2025rolling}, we compute $\Delta^{\text{Quality}}_{\text{Drift}}$ as the absolute difference in aesthetic quality between the first and the last five seconds of each 50-second generated video. Our results demonstrate a clear trend (Fig.~\ref{fig:sink_depth_abl}): as the sink frame size increases, Overall Consistency improves and Aesthetic Quality Drift ($\Delta^{\text{Quality}}_{\text{Drift}}$) decreases. This finding suggests that intermediate frames function as crucial anchors, effectively maintaining both temporal coherence and visual fidelity throughout the long generation process. Consequently, we find that effective attention sinking in Self Forcing~\cite{huang2025self} emerges from deep, extended temporal anchoring, a mechanism that differs from the shallow, initial-frame fixation used in StreamingLLM~\cite{xiao2023efficient}. 

\paragraph{Temporal RoPE Adjustment.}

\label{sec:time-comp}

RoPE~\cite{su2024roformer} is widely used as the positional embedding in video diffusion models, and recent architectures~\cite{wan2025, teng2025magi, yang2024cogvideox, chen2025sana} commonly adopt 3D RoPE, which encodes temporal, height, and width dimensions separately. However, attention sinks in video require the model to attend to past frames, and directly applying 3D RoPE under this setting leads to large temporal discrepancies, where tokens at vastly different timestamps (e.g., $t=1$ vs.\ $t=200$) are forced to attend to each other. This breaks the continuity of video and results in (1) flickering, (2) fidelity degradation, and (3) roll-back, where previously sinked frames are regenerated (a detailed analysis is provided in the Appendix~\ref{sec:deepvs}). To address this, we propose selectively adjusting only the temporal dimensions while preserving the original spatial encoding.

Specifically, we selectively modify the temporal RoPE index by applying a temporal offset to the attention sink's temporal index. This reduces the temporal gap between the attention sink and the remaining tokens, while preserving the spatial indices unchanged.

We divide the key and value caches $K$ and $V$ in the current sliding window into two parts: the sink ($K_{\text{sink}}, V_{\text{sink}}$) for deep sink tokens and the tail ($K_{\text{tail}}, V_{\text{tail}}$) for the rest.
\begin{equation}
\label{eq:win_assem}
{K} \;=\; \big[\,{K}_{\text{sink}}\ \Vert\ {K}_{\text{tail}}\,\big],
\end{equation}
\begin{equation}
{V} \;=\; \big[\,{V}_{\text{sink}}\ \Vert\ {V}_{\text{tail}}\,\big],
\end{equation} where $\big[\cdot\Vert\cdot\big]$ denotes concatenation.

Let $s_\text{tail}$ denote the first frame index of the tail and let $s_{\text{sink}}$ be the last frame index of the deep sink.

We then define $\Delta_\text{sink}$, which is the temporal gap between $s_\text{tail}$ and $s_\text{sink}$, as follows:
\begin{equation}
\Delta_\text{sink} \;=\; s_\text{tail} - s_{\text{sink}}.
\end{equation}

We apply $\Delta_{\text{sink}}$ to $K_{\text{sink}}^{(\text{time})}$, which is temporal component of $K_{\text{sink}}$, using the RoPE temporal frequency vector $\boldsymbol{\omega}_t$:
\begin{equation}
\label{eq:time_comp}
{K}_{\text{sink}}^{(\text{time})}
\;\leftarrow\;
{K}_{\text{sink}}^{(\text{time})}
\odot \exp\!\big(i\,\boldsymbol\omega_t\,\Delta_\text{sink}\big),
\end{equation}
where $i$ is the imaginary unit, $\odot$ denotes element-wise multiplication. This further rotates $K_\text{sink}$ to align the relative temporal positions of sink and tail tokens.

\subsection{Participative Compression}
\label{sec:participated}
\paragraph{Motivation.}

While Deep Sink effectively mitigates fidelity degradation compared to the baseline Self Forcing~\cite{huang2025self}, it alone cannot fully alleviate quality degradation in minute-long video generation. When extrapolating from 5-second training clips to sequences more than 12× longer, a critical issue emerges: \textit{degeneration}, where visual fidelity and overall quality progressively deteriorate. This phenomenon is well-documented in autoregressive long-context generation~\cite{Holtzman2019TheCC,Ghadia2025DialogueWL}: when generating beyond training length, indiscriminate token retention causes attention to dilute across both relevant and irrelevant context, introducing compounding noise. Beyond its training distribution, the growing KV cache retains increasingly irrelevant tokens, further diluting attention.

Recent analysis of video diffusion models reveals that attention concentrates on a small subset of semantically critical tokens, with the majority contributing minimally to generation~\cite{yang2025sparse}—suggesting that pruning low-attention tokens can substantially reduce computation with limited impact on quality. Building on this insight and importance-aware compression~\cite{zhang2023h2o,li2024snapkv,Ghadia2025DialogueWL}, we propose \textbf{Participative Compression}, which dynamically identifies and retains only contextually relevant tokens while pruning those that could contribute to attention dilution and error accumulation.

\paragraph{Overview.}
\label{sec:opc}
Self Forcing~\cite{huang2025self} implements the rolling KV cache by evicting the earliest frame when the cache is filled. 
In comparison, Participative Compression (PC) operates at the token level, selectively removing redundant tokens by ranking them according to their aggregated attention scores from recent frames, rather than using a simple FIFO (First-In, First-Out) policy as illustrated in Fig~\ref{fig:abstract_comp}.

PC introduces two key hyperparameters: (1) $Budget$ ($N$), the target number of tokens to retain after compression, and (2) $Recent$ ($R$), the number of tokens from the most recent frames that are excluded from compression, in addition to the $S$ sink frame tokens that are always preserved. 
PC is applied when the sliding window reaches maximum length $M$ tokens, compressing the cache to size $N \leq M$. The compression operates on  $K_{\text{cand}}, V_{\text{cand}}$, which contain all tokens except those from the first $S$ and the most recent $R$.

\begin{itemize}[leftmargin=*,topsep=3pt,itemsep=1pt]

    \item \textbf{Recent}: $K_{\text{rct}}, V_{\text{rct}}$ containing tokens from the most recent $R$ frames, excluded from compression to preserve local coherence.
    \item \textbf{Candidate}: $K_{\text{cand}}, V_{\text{cand}}$ containing all intermediate tokens between the Sink and Recent, subject to compression.
\end{itemize}

For each token in $K_{\text{cand}}, V_{\text{cand}}$, PC computes an importance score by summing its attention weights from all recent $R$ frames—tokens frequently attended to are deemed critical for maintaining temporal coherence.
PC then selects the top $C = N - R - S$ tokens with the highest importance scores to form $K_{\text{top}}, V_{\text{top}}$. The final KV cache contains $N$ tokens: $S$ sink tokens, $C$ compressed tokens, and $R$ recent tokens.

\paragraph{Top-$C$ Selection.}
\label{sec:topc}
PC selectively retains the $C$ most important candidate tokens based on their relevance to current generation, evicting those not selected by the $\operatorname{Top-C}$ operator.
To determine which tokens to retain, PC computes attention scores between the recent queries ($Q_\text{rct}$) and candidate keys ($K_\text{cand}$). We aggregate these scores across all recent queries by summing along the query dimension, producing a unified importance score $\phi_j$ for each candidate key:
\begin{equation}
\label{eq:fused-mean}
\textstyle
\phi_j = \sum_{r=1}^{R} \mathbf{q}_{r}^{\top}\mathbf{k}_j,
\end{equation}
where $j$ indexes the candidate keys, $\mathbf{q}_r$ denotes the $r$-th query in $Q_\text{rct}$, and $\mathbf{k}_j$ denotes the $j$-th key in $K_\text{cand}$. Higher $\phi_j$ indicates higher importance for current generation. We then form the importance vector $\boldsymbol{\phi} = [\phi_1, \phi_2, \ldots, \phi_{|K_\text{cand}|}]$ and select the Top-$C$ tokens with the highest scores:
\begin{equation}
\label{eq:topc}
K_\text{top} = \operatorname{Top-C}(\boldsymbol{\phi}).
\end{equation}

Finally, the compressed cache is formed by concatenating the preserved components in temporal order:
\begin{equation}
\label{eq:keep}
K_{\text{compressed}} = \big[\,K_{\text{sink}}\ \Vert\ K_{\text{top}}\ \Vert\ K_{\text{rct}}\,\big],
\end{equation}
where $K_{\text{rct}}$ contain keys from the first $S$ and most recent $R$, respectively. Values ($V_\text{top}$) are processed identically. 
This yields a compact cache structure combining long-term initial context (Sink), selectively important intermediate tokens (Top-$C$), and fresh recent context (Recent), all within a fixed budget of $N$.

\begin{table*}[t]
\centering
\scriptsize
\setlength{\tabcolsep}{3pt}
\renewcommand{\arraystretch}{1.08}

\caption{{Quantitative comparison on long video generation.} We evaluate Deep Forcing against open-source autoregressive video diffusion generation baselines on 30-second and 60-second videos across multiple quality metrics on VBench-Long~\cite{huang2024vbench}. 
}
\label{tab:main}
\begin{threeparttable}
\begin{tabularx}{\textwidth}{
  @{}
  >{\raggedright\arraybackslash}p{0.16\textwidth}
  >{\centering\arraybackslash}p{0.10\textwidth}
  *{7}{>{\centering\arraybackslash}X}
  @{}
}
\toprule
\textbf{Model} &
\makecell{\textbf{Throughput}\\(FPS)~$\uparrow$} &
\makecell{\textbf{Dynamic}\\\textbf{Degree}~$\uparrow$} &
\makecell{\textbf{Motion}\\\textbf{Smoothness}~$\uparrow$} &
\makecell{\textbf{Overall}\\\textbf{Consistency}~$\uparrow$} &
\makecell{\textbf{Imaging}\\\textbf{Quality}~$\uparrow$} &
\makecell{\textbf{Aesthetic}\\\textbf{Quality}~$\uparrow$} &
\makecell{\textbf{Subject}\\\textbf{Consistency}~$\uparrow$} &
\makecell{\textbf{Background}\\\textbf{Consistency}~$\uparrow$} \\
\midrule

\rowcolor{black!5}
\multicolumn{9}{l}{
  \makebox[0pt][l]{\textit{Trained with Attention Sink}}%
  \hspace{0.47\textwidth}{\textbf{ 30 seconds}}%
} \\
\cmidrule(lr){2-9}
Rolling Forcing~\cite{liu2025rolling}  & 15.79 & 30.71 & 98.75 & 20.99 & 70.58 & 60.24 & 98.12 & 96.91 \\
LongLive~\cite{yang2025longlive}        & 18.16 & 45.55 & 98.76 & 20.16 & 69.07 & 61.51 & 97.97 &  96.83 \\
\rowcolor{black!5}\multicolumn{9}{l}{\textit{Trained \textbf{without} Attention Sink}} \\
CausVid~\cite{yin2025slow}         & 15.78 &  47.21 & 98.08 & 19.15 & 66.36 & 59.77 & 97.92 & 96.77 \\
Self Forcing~\cite{huang2025self}     & 15.78 & 36.62 & 98.63 & 20.50 & 68.58 & 59.44 & 97.34 & 96.47 \\
\rowcolor{black!2} \textbf{Deep Forcing (Ours)} & 15.75 & 57.56 & 98.27 & 20.54 & 69.31 & 60.68 & 97.34 & 96.48 \\

\midrule  

\rowcolor{black!5}
\multicolumn{9}{l}{
  \makebox[0pt][l]{\textit{Trained with Attention Sink}}%
  \hspace{0.47\textwidth}{\textbf{ 60 seconds}}%
} \\
\cmidrule(lr){2-9}
Rolling Forcing~\cite{liu2025rolling}  & 15.79 & 31.35 & 98.69 & 20.64 & 70.25 & 59.75 & 97.97 & 96.76 \\
LongLive~\cite{yang2025longlive}        & 18.16 & 43.49 & 98.75 & 20.29 & 69.11 & 61.29 & 97.85 & 96.74 \\
\rowcolor{black!5}\multicolumn{9}{l}{\textit{Trained \textbf{without} Attention Sink}} \\
CausVid~\cite{yin2025slow}         & 15.78 & 46.44 & 98.09 & 18.78 & 65.84 & 59.42 & 97.81 & 96.75 \\
Self Forcing~\cite{huang2025self}    & 15.78 & 31.98 & 98.21 & 18.63 & 66.33 & 56.45 & 96.82 & 96.31 \\
\rowcolor{black!2} \textbf{Deep Forcing (Ours)} & 15.75 & 57.19 & 98.23 & 20.38 & 69.27 & 59.86 & 96.96 & 96.32 \\

\bottomrule
\end{tabularx}
\end{threeparttable}
\end{table*}

\begin{algorithm}[t]
\caption{Participative Compression with Deep Sink}
\label{alg:participated}
\small
\begin{algorithmic}[1]
\setlength{\itemsep}{2pt}

\Require KV cache $[K, V]$ of size $M$; Sink size $S$; Recent $R$; Top-C capacity $C$; Timestep $t$; first time step $T$;
\If{$M \geq \text{MAX\_WINDOW\_LENGTH}$ \textbf{and} $t = T$}

    \State \textcolor{gray}{// Partition cache into three regions}
    \State $\mathcal{I}_{\text{sink}} \gets [0, S)$ \Comment{First $S$ frames}
    \State $\mathcal{I}_{\text{rct}} \gets [M-R, M)$ \Comment{Last $R$ frames}
    \State $\mathcal{I}_{\text{cand}} \gets [S, M-R)$ \Comment{Candidate tokens/frames}
    
    \If{$|\mathcal{I}_{\text{cand}}| > 0$ \textbf{and} $C > 0$}
        \State \textcolor{gray}{// Compute importance scores (Eq.~\ref{eq:fused-mean})}
        \State $Q_{\text{rct}} \gets Q[\mathcal{I}_{\text{rct}}]$ \Comment{Recent queries}
        \State $K_{\text{cand}} \gets K[\mathcal{I}_{\text{cand}}]$ \Comment{Candidate keys}
        \For{$j = 1$ to $|\mathcal{I}_{\text{cand}}|$}
            \State $\phi_j \gets \sum_{r=1}^{R} \mathbf{q}_r^\top \mathbf{k}_j$ \Comment{Aggregate attention}
        \EndFor
        
        \State \textcolor{gray}{// Select top-C tokens (Eq.~\ref{eq:topc})}
        \State $\boldsymbol{\phi} \gets [\phi_1, \phi_2, \ldots, \phi_{|\mathcal{I}_{\text{cand}}|}]$
        \State $\mathcal{I}_{\text{top}} \gets \textsc{TopC}(\boldsymbol{\phi})$ \Comment{Select $C$ highest}
        
        \State \textcolor{gray}{// Temporal RoPE Unification (Section~\ref{sec:unify})}
        \State $\Delta_{\text{top}} \gets s^{\text{top}} - s^{\text{top}}_{\text{base}}$ 
        \State $K_{\text{top}}^{(\text{time})} \gets K_{\text{top}}^{(\text{time})} \odot \exp(i\boldsymbol{\omega}_t \Delta_{\text{top}})$
    \Else
        \State $\mathcal{I}_{\text{top}} \gets \varnothing$
    \EndIf
    
    \State \textcolor{gray}{// Assemble compressed cache (Eq.~\ref{eq:keep})}
    \State $K_{\text{compressed}} \gets [K_{\text{sink}}\ \Vert\ K_{\text{top}}\ \Vert\ K_{\text{rct}}]$
    \State $V_{\text{compressed}} \gets [V_{\text{sink}}\ \Vert\ V_{\text{top}}\ \Vert\ V_{\text{rct}}]$
    \State \Return $K_{\text{compressed}}, V_{\text{compressed}}$
\Else
    \State \Return $K, V$ \Comment{No compression}
\EndIf
\end{algorithmic}
\end{algorithm}

\paragraph{Temporal RoPE Unification.}
\label{sec:unify}

After selecting the Top-$C$ tokens, we apply RoPE adjustment to maintain temporal dimension consistency, following the same approach as Deep Sink (Section~\ref{sec:deep-sink}). 
We adjust only the temporal dimension of the Top-$C$ keys' RoPE while preserving their spatial information intact.

Let $s^{\text{top}}$ denote the desired absolute temporal position where the Top-$C$ block should be aligned, and let $s^{\text{top}}_{\text{base}}$ represent the current temporal position of each cached Top-$C$ key. We compute the temporal adjustment:
\begin{equation}
\Delta_{\text{top}} \;=\; s^{\text{top}} - s^{\text{top}}_{\text{base}}.
\end{equation}

This temporal shift $\Delta_{\text{top}}$ is then applied to $K_{\text{top}}^{(\text{time})}$, which is temporal component of $K_{\text{top}}$, re-aligning each Top-$C$ key using the complex phase rotation defined by the RoPE temporal frequencies $\boldsymbol\omega_t$:
\begin{equation}
\label{eq:topc_rope}
{K}^{(\text{time})}_{\text{top}}
\;\leftarrow\;
{K}^{(\text{time})}_{\text{top}}
\odot \exp\!\big(i\,\boldsymbol\omega_t\,\Delta_{\text{top}}\big).
\end{equation} where $i$ is the imaginary unit, and $\odot$ denotes element-wise multiplication.

This rotation adjusts the temporal positioning of $K_{\text{top}}$ to create a continuous temporal sequence across all three cache components (Sink, Top-$C$, Recent), preventing temporal discontinuities that would otherwise cause fidelity degradation, flickering, and roll-back artifacts as demonstrated in Appendix~\ref{sec:deepvs}.

\paragraph{Efficiency.}
\label{sec:effic}
The complexity of Participative Compression (PC) might initially suggest a significant computational overhead. However, its computational burden is minimized due to its sparse activation criteria. $\text{PC}$ is only engaged under two specific conditions: When the sliding context window is completely filled, and during the first diffusion time step ($t=T$). Even though the Top-$C$ selection mechanism involves gathering and sorting tokens, our efficiency analysis justifies this operation in the Appendix~\ref{sec:fps}.

\section{Experiments}

\subsection{Experimental settings}
\paragraph{Implementation details.}
We implement chunk-wise Self Forcing~\cite{huang2025self} as our base model. We evaluate both quantitatively and qualitatively using Deep Sink (DS) combined with Participative Compression (PC). 
We set the hyperparameters for Deep Sink and Participative Compression as follows: sink size $S=10$, budget $N=16$, and recent $R=4$. We compare against baseline autoregressive video diffusion models including CausVid~\cite{yin2025slow}, Self Forcing~\cite{huang2025self}, Rolling Forcing~\cite{liu2025rolling}, and LongLive~\cite{yang2025longlive}.

\paragraph{Evaluation.}
We conduct evaluations under two settings. First, we evaluate long video generation on VBench-Long~\cite{huang2024vbench} using 128 prompts from MovieGen~\cite{polyak2024movie}, following the same prompt selection protocol as Self Forcing++~\citep{cui2025self}. Each prompt is refined using Qwen/Qwen2.5-7B-Instruct~\cite{yang2025qwen3} following Self Forcing~\cite{huang2025self}. 
Second, we conduct a user preference study to evaluate color consistency, dynamic motion, subject consistency, and overall quality.
Additional implementation details are provided in the Appendix~\ref{sec:userstudy}.  
Third, we evaluate visual stability using the state-of-the-art vision-language model (VLM) Gemini 2.5-Pro~\cite{comanici2025gemini}, following the protocol of Self Forcing++~\cite{cui2025self}.

\begin{figure*}[t]
  \centering
\includegraphics[width=1.0\textwidth,clip]{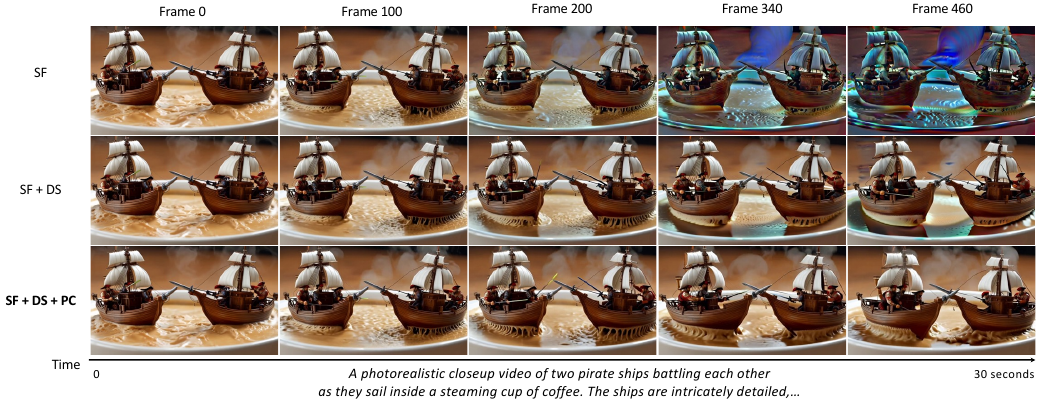}
    \caption{\textbf{Qualitative ablation results over 30-second generation:} Comparison of Self Forcing (SF)~\cite{huang2025self}, SF with Deep Sink (SF+DS), and SF with both Deep Sink and Participative Compression (Deep Forcing). Baseline SF exhibits severe color drift. SF+DS improves stability but shows residual artifacts. Deep Forcing maintains consistent visual quality.}\vspace{-10pt}
  \label{fig:ablation_qual}
\end{figure*}

\subsection{Results in Long Video Generation}
\paragraph{Quantitative results.}
Our quantitative results are presented in Table~\ref{tab:main}. Despite being a training-free method built upon Self Forcing, which was not trained for long video generation, our approach achieves performance comparable to methods explicitly distilled or trained for long videos~\cite{yang2025longlive, liu2025rolling}. As shown in Table~\ref{tab:main}, we achieve superior performance in overall consistency and imaging quality compared to LongLive~\cite{yang2025longlive}, and better aesthetic quality than Rolling Forcing~\cite{liu2025rolling}.

Notably, our method also excels in dynamic degree, producing more dynamic motions than trained methods~\cite{liu2025rolling, yang2025longlive}, despite not being explicitly optimized for this aspect. We attribute this to our training-free approach, which avoids the potential motion constraints introduced when models are explicitly trained to anchor with attention sinks.

\paragraph{Qualitative results.}

\begin{figure*}[t!]
  \centering
  \includegraphics[width=0.95\textwidth,trim={0 0.5cm 0 0.5cm},clip]{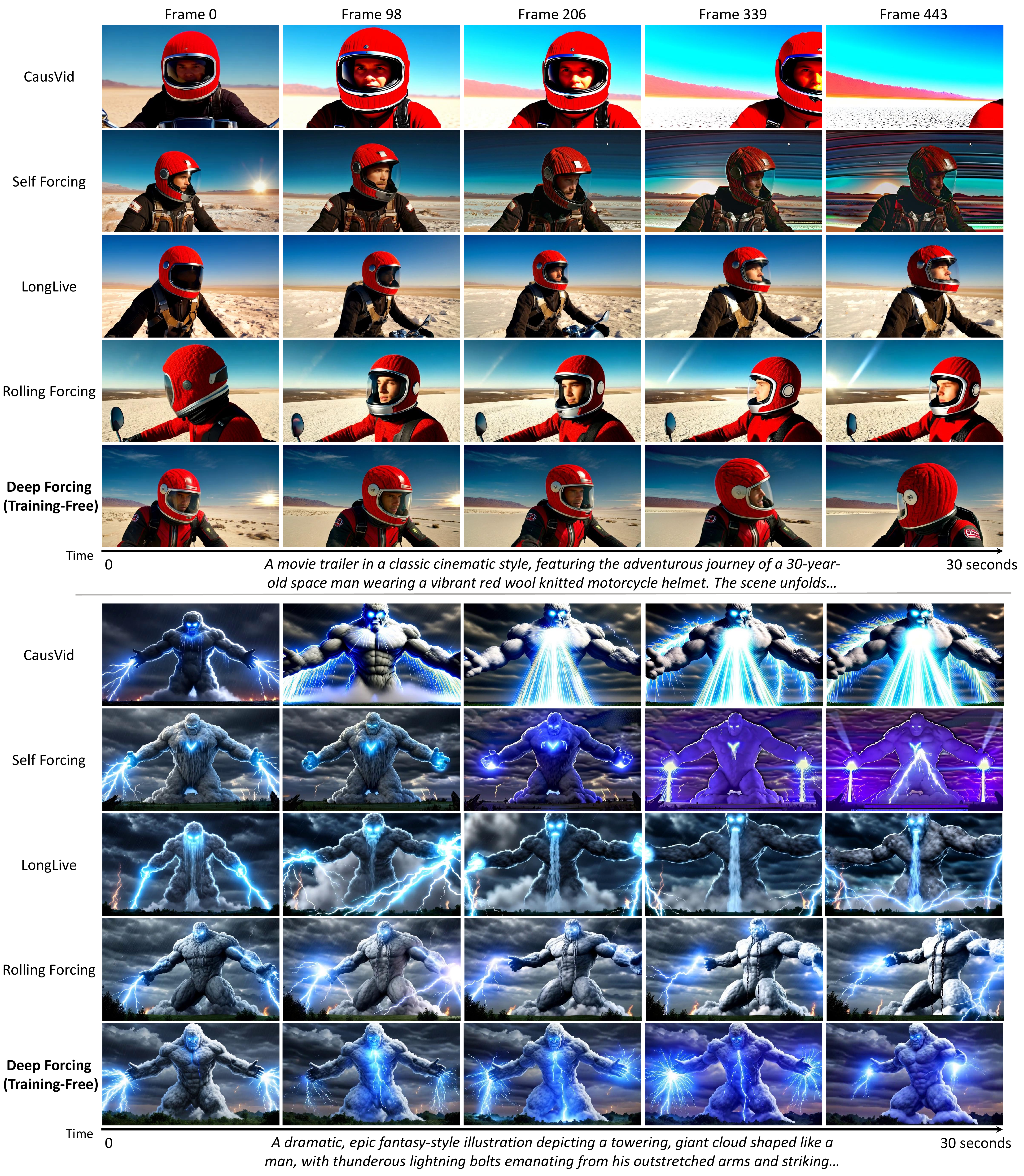}
  \caption{\textbf{Qualitative results on 30-second videos.} Frame-by-frame comparison across different methods for two representative prompts. Deep Forcing (training-free) achieves temporal consistency and visual quality comparable to training-based baselines (CausVid~\cite{yin2025slow}, Self Forcing~\cite{huang2025self}, LongLive~\cite{yang2025longlive}, Rolling Forcing~\cite{liu2025rolling}) while generating more dynamic content with greater subject consistency.}
  \label{fig:main_qual}
\end{figure*}

The qualitative results in Figure~\ref{fig:main_qual} demonstrate strong visual quality, with our training-free method producing high-fidelity frames comparable to or better than training-based baselines. The results visually confirm our quantitative performance, where we achieve competitive scores without any fine-tuning. Notably, our videos exhibit more dynamic motion in both camera and subject movement, yielding more visually expressive results compared to existing approaches.

Although subject consistency is lower in VBench-Long metrics, the bottom example in Figure~\ref{fig:main_qual} demonstrates that our training-free approach maintains better overall quality with limited degradation compared to training-based methods. Additional qualitative results are provided in Appendix~\ref{sec:supqual}.

\paragraph{User study.}
To further validate these observations, we conducted a user study with 24 participants evaluating multiple aspects of the generated videos. 
The user study was conducted following the Two-Alternative Forced Choice (2AFC) protocol, where users are instructed to choose which is better between two videos (from Deep Forcing and a baseline), in regard to color consistency, dynamic motion, subject consistency, and overall quality. 
As shown in Table~\ref{tab:user_study}, participants demonstrated a clear preference for Deep Forcing over the baselines across all evaluated aspects. 
This includes a high preference in terms of subject consistency, highlighting Deep Forcing's ability to retain the subject with minimal identity drift throughout the video.
These corroborate our qualitative assessment that perceptual quality remains high despite lower VBench-Long~\cite{huang2024vbench} subject consistency scores. 

\begin{table}[t]
  \caption{\textbf{User study results.} Values represent \textit{percentage of votes favoring our Deep Forcing} over the baselines.}
  \label{tab:user_study}
  \centering
  \scriptsize
  \setlength{\tabcolsep}{4pt}
  \renewcommand{\arraystretch}{1.10}
  \begin{tabular}{@{}lcccc@{}}
    \toprule
    Method & Color Cons. & Dyn. Motion & Subject Cons. & Overall Quality \\
    \midrule
    CausVid             & 98.9\% & 95.8\% & 96.8\% & 100\% \\
    Self Forcing         & 85.9\% & 86.9\% & 84.8\% & 87.9\% \\
    LongLive          & 71.2\% & 83.5\% & 72.2\% & 72.2\% \\
    Rolling Forcing      & 76.7\% & 76.7\% & 80.0\%  &78.9\% \\
    \bottomrule
  \end{tabular}
\end{table}

\paragraph{VLM evaluation.}
For further comparison with the baselines, we evaluate visual stability using the state-of-the-art vision–language model (VLM) Gemini 2.5-Pro~\cite{comanici2025gemini}. Following the protocol of Self Forcing++~\cite{cui2025self}, we use the same prompt to ask the VLM to score each 30-second video in terms of exposure stability and degradation. Then we normalize the resulting scores 100. As shown in Tab.~\ref{tab:vlm_evaluation}, our training-free method achieves visual stability comparable to that of recent methods~\cite{liu2025rolling,yang2025longlive}.

\label{tab:vlm_evaluatio}
\begin{table}[t]
  \caption{\textbf{Visual stability compared with the baselines.}
  Methods are additionally categorized by whether they are trained with an attention sink.}
  \label{tab:vlm_evaluation}
  \centering
  \scriptsize
  \setlength{\tabcolsep}{5pt}
  \renewcommand{\arraystretch}{1.10}
  \begin{tabular}{@{}lcc@{}}
    \toprule
    Method & Attention Sink Training & Visual Stability \\
    \midrule
    CausVid~\cite{yin2025slow}              & No  & 42.84 \\
    Self Forcing~\cite{huang2025self}       & No  & 43.94 \\
    Rolling Forcing~\cite{liu2025rolling}   & Yes & 72.6 \\
    LongLive~\cite{yang2025longlive}        & Yes & 78.58 \\
    \midrule
    \textbf{Deep Forcing (Ours)}            & No  & 75.44 \\
    \bottomrule
  \end{tabular}
\end{table}

\subsection{Ablation studies}
\label{sec:abl}
We conducted ablation studies to evaluate the contributions of each method. We measure relevant VBench-Long metrics on 30 second videos. 

\paragraph{Effect of Deep Sink \& Participative Compression.}
We evaluate three variants: naive Self-Forcing~\cite{huang2025self}, Self-Forcing with only Deep Sink with sink length $S$ = 10 frames, and Self-Forcing with both Deep Sink ($S$ = 10 frames) and Participative Compression ($N$ = 16, $R$ = 4). As shown in Table~\ref{tab:DS_PC_abl}, Deep Forcing demonstrates progressive improvements in dynamic degree, overall consistency, and image quality as components are added. 
Notably, dynamic degree improves substantially through the Deep Forcing framework, enabling the generation of significantly more dynamic scenes compared to baseline methods.

While image quality shows a slight decrease at 30 seconds, we have already demonstrated Deep Sink's positive impact on 50-second videos in Section~\ref{sec:deep-sink}.

\paragraph{Ablation Visualization.}
Figure~\ref{fig:ablation_qual} visualizes our ablation study results. When generating long videos with Self Forcing (SF) alone (top row), error accumulation leads to severe fidelity degradation and visual quality deteriorates as colors drift toward over-saturation. Adding Deep Sink (SF + DS) substantially reduces fidelity degradation and maintains more consistent colors. However, subtle artifacts persist at frame 460, including slight color shift in the coffee and texture blur in the ship details. When both Deep Sink and Participative Compression are applied (\textbf{Deep Forcing}), noticeable degradation is effectively eliminated. This validates that our complete framework successfully mitigates long-horizon error accumulation while preserving both overall visual quality and fine-grained details.

\begin{table}[t]
  \caption{\textbf{Ablation on our components}: Deep Sink (DS) and Participative Compression (PC).}
  \label{tab:DS_PC_abl}
  \centering
  \scriptsize
  \setlength{\tabcolsep}{4pt}
  \renewcommand{\arraystretch}{1.10}
  \begin{tabular}{@{}lccc@{}}
    \toprule
    Method & Dynamic Degree & Overall Consistency & Image Quality \\
    \midrule
    SF~\cite{huang2025self} (Baseline)            & 36.62 & 20.50 & 68.58 \\
    SF~\cite{huang2025self} + DS         & 48.58 & 20.54 & 68.54 \\
    SF~\cite{huang2025self} + DS + PC (\textbf{Ours})    & 57.56 & 20.54 & 69.31 \\
    \bottomrule
  \end{tabular}
\end{table}

\paragraph{Top-C Visualization.}
\begin{figure}[t]
  \centering
   \includegraphics[width=1.0\linewidth]{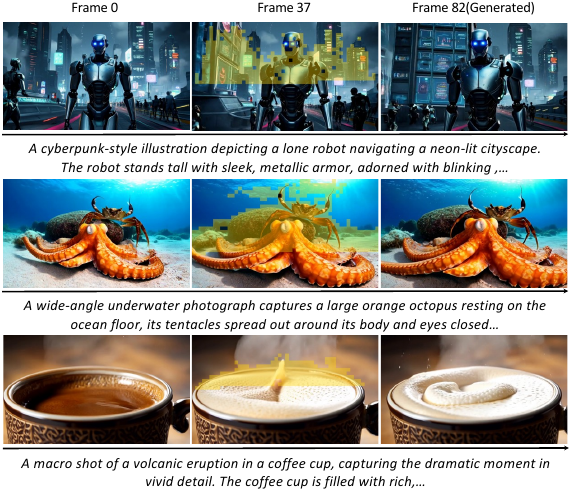}

      \caption{\textbf{Visualization of Top-$C$ token selection.} For each example, Frame 37 \textbf{(middle)} shows the Top-$C$ tokens selected for generating Frame 82 \textbf{(right)}. {Yellow} highlights indicate the spatial locations of tokens chosen as Top-$C$. Our method effectively identifies and preserves regions that are critical for maintaining contextual coherence during subsequent generation.}
   \label{fig:topc_vis_main}
\end{figure}

Figure~\ref{fig:topc_vis_main} visualizes a subset of Top-$C$ tokens selected during the first rolling step, spatially aligned to their positions in Frame 37 when generating Frame 82. The yellow highlighted regions indicate the spatial positions of tokens selected as Top-$C$ within the frame. These highlighted regions reveal  semantic alignment with contextually important content: the robot's body and background architecture, the octopus tentacles and crab, and the circular coffee cup structure. This demonstrates that our method identifies and retains semantically salient regions critical for maintaining contextual coherence in subsequent generation.

\section{Conclusion}
\label{sec:conclusion}
We introduced Deep Forcing, a training-free approach for autoregressive long video generation that effectively mitigates error accumulation through two key components: \textbf{Deep Sink} and \textbf{Participative Compression}. Our method achieves state-of-the-art performance on VBench-Long, user studies, and VLM evaluation without any fine-tuning, even surpassing training-based methods. By exploiting the inherent deep attention sink behavior in pre-trained Self Forcing, we enable minute-long video generation while preserving both visual fidelity and motion dynamics. This training-free paradigm offers a practical and efficient solution for long video generation with autoregressive video diffusion.

\paragraph{Limitations and Future Works.} \label{sec:limitation} While our training-free approach substantially improves long-horizon stability, several limitations remain. Operating at inference time on a frozen backbone, our method is constrained by the pretrained model's capacity and biases. Additionally, our approach lacks explicit long-term memory, potentially causing gradual drift in extremely long sequences with repeated occlusions. Future work could integrate hierarchical memory modules and extend to broader video generation settings.

\clearpage

\clearpage
{
    \small
    \bibliographystyle{ieeenat_fullname}
    \bibliography{main}
}

\clearpage
\setcounter{page}{1}

\setcounter{section}{0} 
\renewcommand{\thesection}{\Alph{section}}
\section*{\Large Appendix}

\begin{figure*}[t]
  \centering
\includegraphics[width=1.0\textwidth,clip]{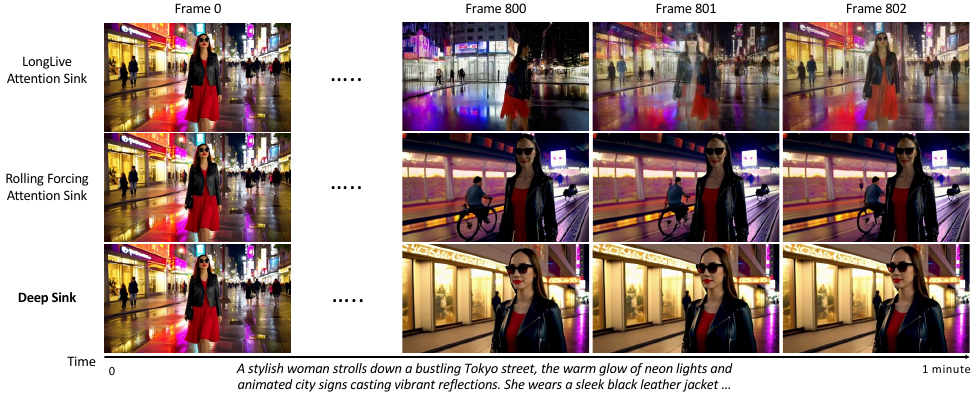}
  \caption{\textbf{Qualitative results on different Attention Sink.} The result shows that Deep Sink substantially outperforms both LongLive-style~\cite{yang2025longlive} and Rolling Forcing-style~\cite{liu2025rolling} attention sinks.}\vspace{-10pt}
  \label{fig:Deep_Forcing_suppl1_qual}
\end{figure*}

In this appendix, we provide comprehensive details, including: 
\begin{itemize}
\item Comparison with other sink mechanisms~\cite{yang2025longlive, liu2025rolling}\\ (Appendix~\ref{sec:deepvs})
\item Qualitative results on different sink sizes (Appendix~\ref{sec:sink_abl_qual})
\item Participative Compression details (Appendix~\ref{sec:PC_detail})
\item Denoising query is not just noise (Appendix~\ref{sec:noisevsquery})
\item FPS measurements (Appendix~\ref{sec:fps})
\item More qualitative results (Appendix~\ref{sec:supqual})
\item User study protocol (Appendix~\ref{sec:userstudy})
\item Additional attention visualization (Appendix~\ref{sec:additional_attn_vis})
\end{itemize}

\section{The Tale of Three Sinks}
\label{sec:deepvs}
Concurrent works~\cite{yang2025longlive, liu2025rolling} propose different attention sink mechanisms for Self Forcing-like architectures, typically with model training or distillation. In this section, we compare these approaches in a training-free setting to evaluate their effectiveness when applied directly to pretrained Self Forcing model.

Specifically, we compare three attention sink strategies: LongLive~\cite{yang2025longlive}, Rolling Forcing~\cite{liu2025rolling}, and Ours. LongLive applies attention sinks to the first \textbf{3 frames} without RoPE adjustment. Rolling Forcing also uses \textbf{3 frame} sinks but incorporates (1) storing raw keys and (2) dynamically re-applying RoPE when rolling occurs. Qualitative comparisons are presented in Figure~\ref{fig:Deep_Forcing_suppl1_qual}.

Figure~\ref{fig:Deep_Forcing_suppl1_qual} shows that our method achieves substantially better results. The LongLive attention sink, which does not adjust RoPE, exhibits progressive failure modes: ~\textbf{fidelity degradation} appears at frame 800, followed by ~\textbf{flickering} artifacts at frame 801, and culminating in ~\textbf{roll-back} behavior at frame 802 where the generation reverts to early sinked frames. These issues also occur in LongLive~\cite{yang2025longlive}, which was explicitly trained on long videos using this attention sink mechanism (Fig.~\ref{fig:teaser}).

Although Rolling Forcing attention sink~\cite{liu2025rolling} employs Dynamic RoPE, which reapplies positional encodings to the entire cached key set, it still exhibits severe fidelity degradation at frames 800--801.

This comparison demonstrates that both deep sink and RoPE adjustment are essential for long video generation.

\section{Qualitative Results on Different Sink Size}
\label{sec:sink_abl_qual}
\begin{figure*}[t]
  \centering
  \includegraphics[width=1.0\textwidth]{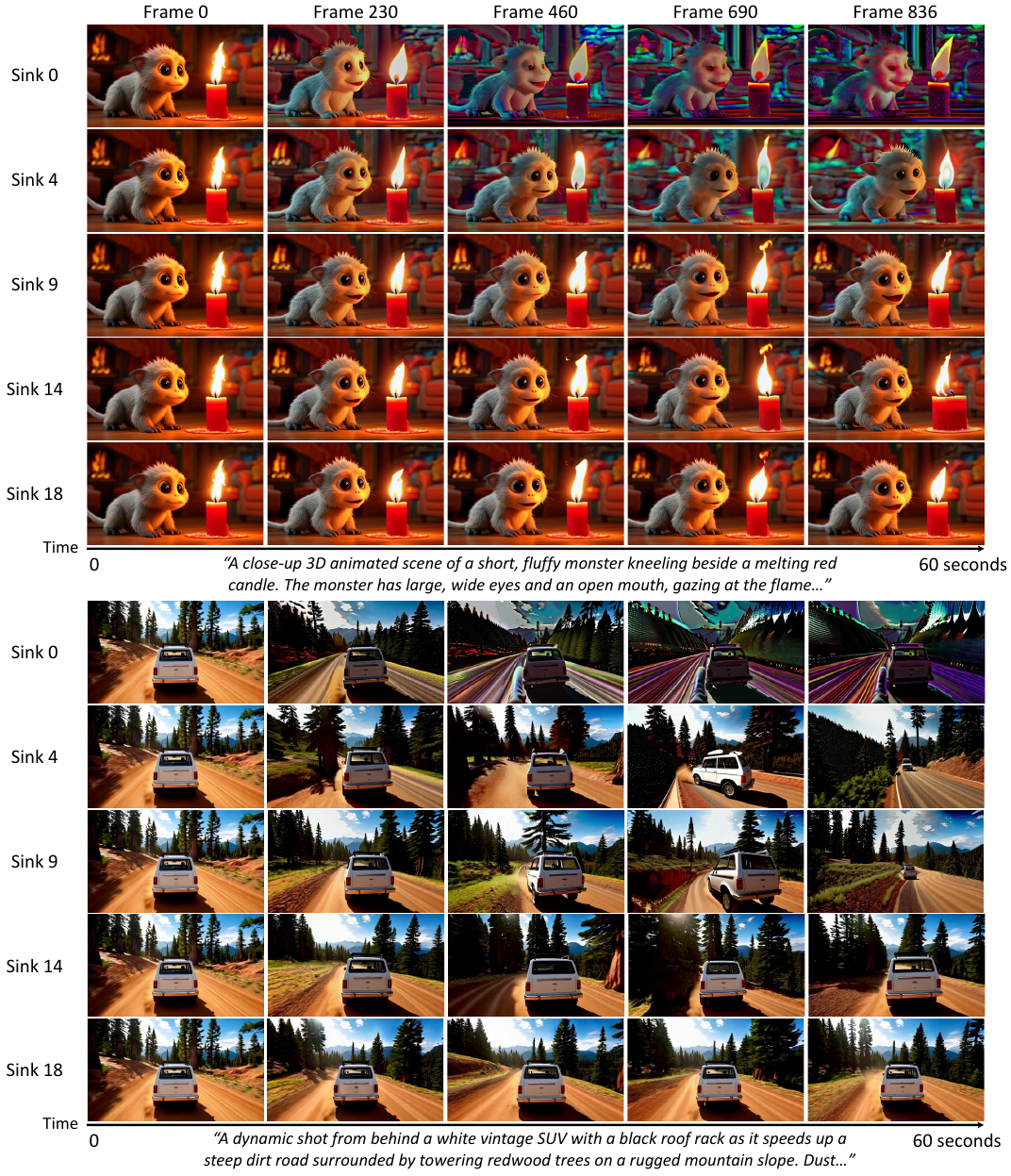}
  \caption{\textbf{Qualitative comparison of different sink sizes on 60-second videos.} As the sink size decreases, degradation becomes more severe. Once the sink size exceeds 10 frames, degradation is substantially reduced.}\vspace{0pt}
  \label{fig:suppl_sinksize_qual}
\end{figure*}

Figure~\ref{fig:suppl_sinksize_qual} presents qualitative comparisons across sink sizes ranging from 0 to 18 frames, as analyzed in Section~\ref{sec:deep-sink}. We evaluate two diverse prompts on 60-second generation.

With no attention sink (Sink 0), severe fidelity degradation emerges rapidly, where the monster's texture deteriorates and colors shift noticeably by frame 230, with complete quality collapse by frame 690. Similarly, the SUV scene exhibits significant fidelity degradation. As the sink size increases to 4 and 9 frames, degradation is progressively reduced but remains visible in fine details.

Once the sink size exceeds 10 frames (Sink 14), fidelity degradation is substantially reduced. However, excessively large sinks (Sink 18) exhibit repetitive generation where early frames are over-preserved.

These results validate our optimal sink range of 10--15 frames (40--60\% of the sliding window). While Deep Sink substantially mitigates degradation, it alone proves insufficient to maintain visual fidelity throughout minute-long generation across diverse scenes, as demonstrated in our extended evaluations (Section~\ref{sec:abl}).

\section{Participative Compression Details}
\label{sec:PC_detail}
In this section, we provide additional details and analysis of Participative Compression (PC) beyond what was presented in Section~\ref{sec:participated}.

Each layer maintains its own KV cache, which undergoes compression as follows:

\begin{equation}
\begin{aligned}
\big[\,K_\text{sink}\ \Vert\ K_{\text{cand}}\ \Vert\ K_\text{rct}\,\big]
&\;\rightarrow\;
\big[\,K_\text{sink}\ \Vert\ K_\text{top}\ \Vert\ K_\text{rct}\,\big], \\[3pt]
\big[\,V_\text{sink}\ \Vert\ V_{\text{cand}}\ \Vert\ V_\text{rct}\,\big]
&\;\rightarrow\;
\big[\,V_\text{sink}\ \Vert\ V_\text{top}\ \Vert\ V_\text{rct}\,\big].
\end{aligned}
\end{equation}

PC compresses only the intermediate tokens between the sink and recent. This design preserves both the initial context ($K_\text{sink}, V_\text{sink}$) and recent context ($K_\text{rct}, V_\text{rct}$) without modification, while compressing only the candidate tokens ($K_{\text{cand}}, V_{\text{cand}}$) to retain the most important visual and contextual information ($K_\text{top}, V_\text{top}$).

This compression occurs independently in each layer. Importantly, PC is applied only at the first diffusion timestep ($t=1000$) when the cache reaches over its maximum window length. The tokens selected at this initial timestep remain fixed throughout subsequent denoising steps.

\begin{figure*}[t]
  \centering
  \includegraphics[width=1.0\textwidth,clip]{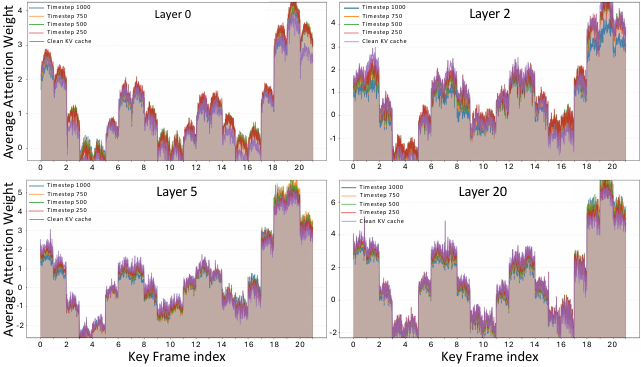}
  \caption{\textbf{Attention weight consistency across diffusion timesteps.} Query-averaged attention weights showing how each key frame is attended when generating the last chunk (frames 19--21) at different denoising timesteps. The consistent attention patterns across timesteps (1000, 750, 500, 250, and the final clean KV cache) demonstrate that Top-$C$ tokens selected at the initial timestep ($t=1000$) remain valid and contextually relevant throughout the entire denoising process.}
  \label{fig:timestep_analysis}
\end{figure*}

Figure~\ref{fig:timestep_analysis} validates this design choice by demonstrating that attention patterns remain consistent across timesteps. The tokens deemed important when generating frames 19--21 exhibit similar importance scores across different diffusion timesteps ($t=1000, 750, 500, 250$), confirming that the Top-$C$ selection at $t=1000$ captures tokens that remain contextually relevant throughout the entire denoising process.

\paragraph{Participative Compression Ablation.} 
As shown in Figure~\ref{fig:architecture} in the main paper, Participative Compression (PC) can leverage both current denoising query tokens and clean past query tokens from previously generated frames to select the Top-$C$ candidates. We evaluate the effect of using each type independently versus combining them together.

Table~\ref{tab:PC_abl} compares these three strategies. When Top-$C$ is selected using only clean past tokens (Only Past), the method achieves an image quality of 68.54 and overall consistency of 20.47. When selection relies solely on currently denoising tokens (Only Denoising), the noisy nature of these queries at the initial timestep ($t=1000$) leads to slightly lower image quality (68.24) and motion smoothness (97.86), likely due to unstable token selection at the initial denoising step.
Combining both query types (Both) achieves the highest scores across all metrics, including motion smoothness, overall consistency, and image quality. The clean past queries appear to provide relatively stable importance estimates, while the current denoising queries help ensure the selected tokens remain relevant to the immediate generation context, suggesting complementary benefits from their combination.

\begin{table}[t]
  \caption{\textbf{Ablation on Participative Compression.}}
  \label{tab:PC_abl}
  \centering
  \scriptsize
  \setlength{\tabcolsep}{4pt}
  \renewcommand{\arraystretch}{1.10}
  \begin{tabular}{@{}lccc@{}}
    \toprule
    Method & Motion Smoothness & Overall Consistency & Image Quality \\
    \midrule
    Only Denoising             & 97.86 & 20.44 & 68.24 \\
    Only Past         & 97.91 & 20.47 & 68.54 \\
    Both                    & 98.27 & 20.54 & 69.31 \\
    \bottomrule
  \end{tabular}
\end{table}

\section{Denoising Query Is Not Just Random Noise}
\label{sec:noisevsquery}
While the denoising queries at the initial timestep ($t=1000$) are inherently noisy, they may still carry meaningful signal for identifying important tokens. Figure~\ref{fig:timestep_analysis} suggests this by showing consistent attention patterns across timesteps, but to more conclusively demonstrate the effectiveness of noisy queries, we directly compare Top-$C$ selection based on denoising queries versus Gaussian random selection. For denoising query-based selection, we compute $QK^T$ using only the currently denoising query tokens, then select the Top-$C$ candidates. For Gaussian random selection, we assign each candidate token a score sampled from $\mathcal{N}(0,1)$ and select the Top-$C$ based on these random scores.

\begin{figure*}[t]
  \centering
  \includegraphics[width=1.0\textwidth,clip]{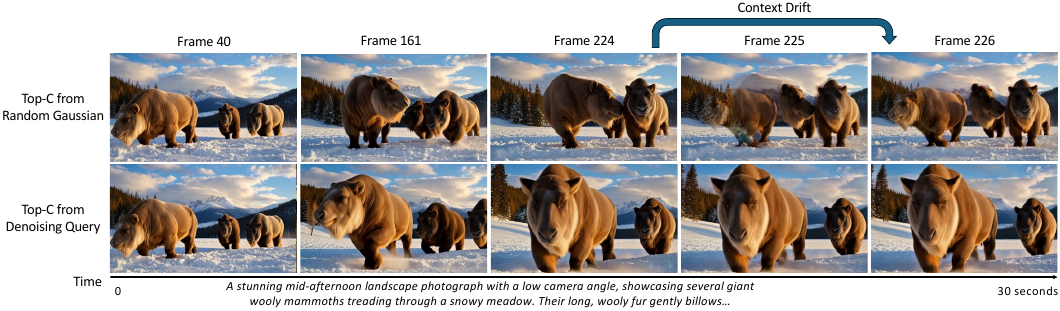}
  \caption{\textbf{Qualitative comparison: Random Top-$C$ vs. Denoising Query Top-$C$.} Gaussian random selection causes severe artifacts during compression - faces abruptly rotate, heads appear floating in mid-air, and random context drift occurs, resulting in incoherent scene transitions. In contrast, denoising query-based selection maintains subject consistency with natural emergent camera movements and preserves contextual coherence throughout the generation.}\vspace{-10pt}
  \label{fig:rand_vs_noise_qual}
\end{figure*}

Figure~\ref{fig:rand_vs_noise_qual} illustrates the stark difference. Random selection exhibits severe scene repetition and context loss, as randomly chosen anchors fail to preserve coherent contextual information. In contrast, denoising query-based selection generates context-aware videos with notably better subject consistency and context.

\begin{figure*}[t]
  \centering
  \includegraphics[width=1.0\textwidth,clip]{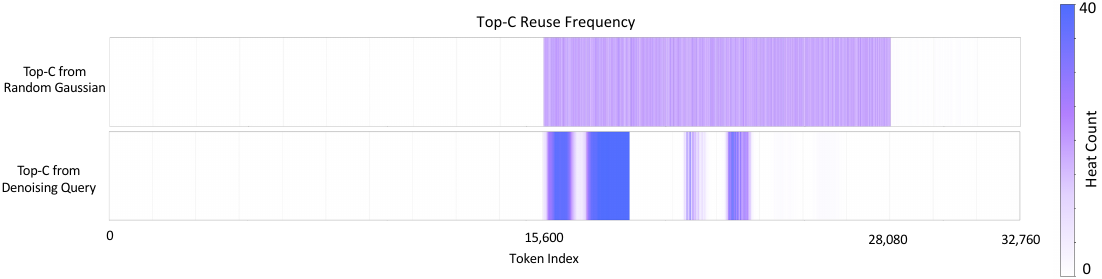}
  \caption{\textbf{Token-wise Top-$C$ selection frequency heatmap during 1-minute generation.}
Color intensity ranges from white (rarely selected) to \textcolor{topcDark}{dark purple} (frequently selected as Top-$C$), indicating how often each token is reused throughout the generation.
The x-axis spans tokens 0–32,760, where 0–15,600 are Deep Sink tokens, 15,600–28,080 are candidates for compression, and 28,080+ are recent tokens. Gaussian random selection (top) distributes selections uniformly across candidate tokens, whereas denoising query-based selection (bottom) concentrates heavily on specific semantically important tokens—particularly those immediately after the sink boundary—that effectively bridge established and newly formed context.}\vspace{-10pt}
  \label{fig:topcfreq}
\end{figure*}

Figure~\ref{fig:topcfreq} further validates this through token selection frequency heatmaps over 1-minute generation, where color intensity (white to \textcolor{topcDark}{dark purple}) indicates selection frequency. The x-axis spans tokens 0--32,760: tokens 0--15,600 are Deep Sink, 15,600--28,080 are compression candidates, and 28,080+ are recent. Gaussian random selection (top) distributes uniformly across candidates, while denoising query-based selection (bottom) concentrates heavily on specific positions, particularly immediately after the sink boundary (15,600).

Notably, these high-frequency positions do not correspond to fixed frames, as tokens are evicted during compression, subsequent tokens shift into these slots. The concentration at positions near 15,600 indicates that these positional slots are consistently selected regardless of frame identity, as they bridge established context (sink) and current generation (recent), serving as semantically important anchors. This positional selectivity demonstrates meaningful contextual relationships rather than arbitrary noise.

We hypothesize this effectiveness stems from: (1) Self Forcing's 4-step distilled diffusion enabling rapid convergence to meaningful attention patterns despite noisy queries at $t=1000$, and (2) per-layer KV caching allowing independent selection of semantically important tokens based on layer-specific contextual relevance.

\section{FPS measurements}
\label{sec:fps}
We evaluate inference throughput on a single NVIDIA H100 GPU when generating 60-second videos.
Table~\ref{tab:DS_PC_abl} demonstrates that Deep Forcing maintains throughput comparable to baseline Self Forcing, achieving 15.75 FPS versus 15.78 FPS. Despite the computational overhead of compression, our method balances two competing factors: (1) compressing from 21 to 16 frames requires additional computation, but (2) generating subsequent frames with only 16 cached frames incurs lower attention costs compared to full attention over 21 frames. 

The latency range after first rolling in Table~\ref{tab:DS_PC_abl} reflects this trade-off. While Deep Forcing exhibits a wider latency range (0.747s to 0.797s) compared to the baseline (0.770s to 0.776s), the average latencies are nearly identical, demonstrating that compression overhead is effectively balanced by reduced attention costs. In practice, throughput oscillates between compression phases (slightly slower) and generation phases (slightly faster) as the cache alternates between 21 and 16 frames. These fluctuations average to nearly identical performance as the baseline, demonstrating that our compression mechanism effectively amortizes its overhead, enabling long-horizon generation with minimal performance penalty.

\begin{table}[t]
  \caption{\textbf{Throughput Comparison on a single H100 GPU. Latency is measured after first rolling.}}
  \label{tab:DS_PC_abl}
  \centering
  \scriptsize
  \setlength{\tabcolsep}{4pt}
  \renewcommand{\arraystretch}{1.10}
  \begin{tabular}{@{}lccc@{}}
    \toprule
    Method & FPS & Latency(Min/Max)  \\
    \midrule
    Self Forcing~\cite{huang2025self}              & 15.78  & 0.770 / 0.776s\\
    \textbf{Deep Forcing (Ours)}        & 15.75  & 0.747 / 0.797s \\
    \bottomrule
  \end{tabular}\vspace{-10pt}
\end{table}

\section{More Qualitative Results}
\label{sec:supqual}
\begin{figure*}[t]
  \centering
  \includegraphics[width=1.0\textwidth]{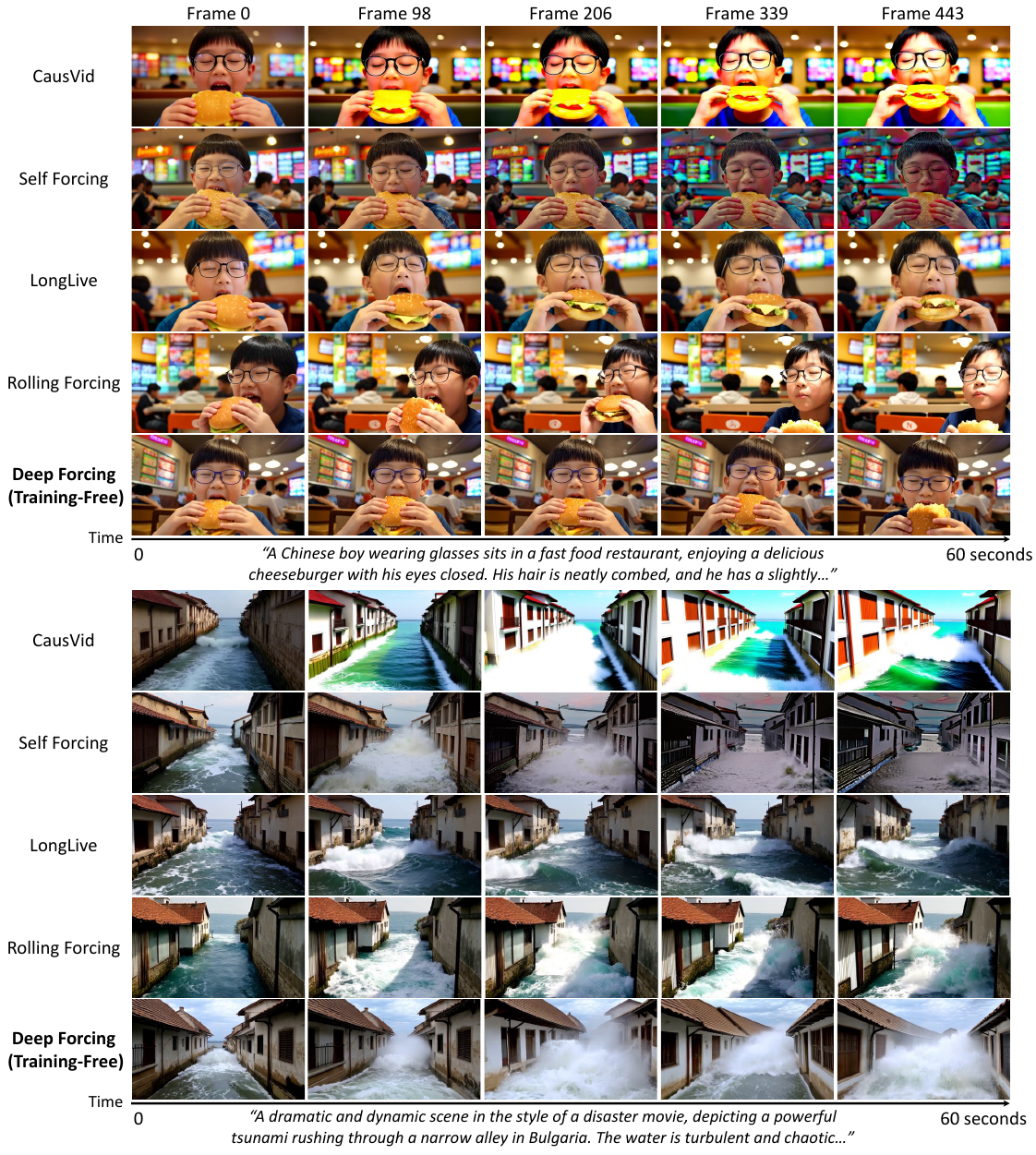}
  \caption{\textbf{Qualitative results on 30-second videos.} Frame-by-frame comparison across different methods for two representative prompts. Deep Forcing (training-free) achieves temporal consistency and visual quality comparable to training-based baselines (CausVid~\cite{yin2025slow}, Self Forcing~\cite{huang2025self}, LongLive~\cite{yang2025longlive}, Rolling Forcing~\cite{liu2025rolling}) while generating more dynamic content with greater subject consistency.}\vspace{10pt}
  \label{fig:suppl_qual1}
\end{figure*}

\begin{figure*}[t]
  \centering
  \includegraphics[width=1.0\textwidth]{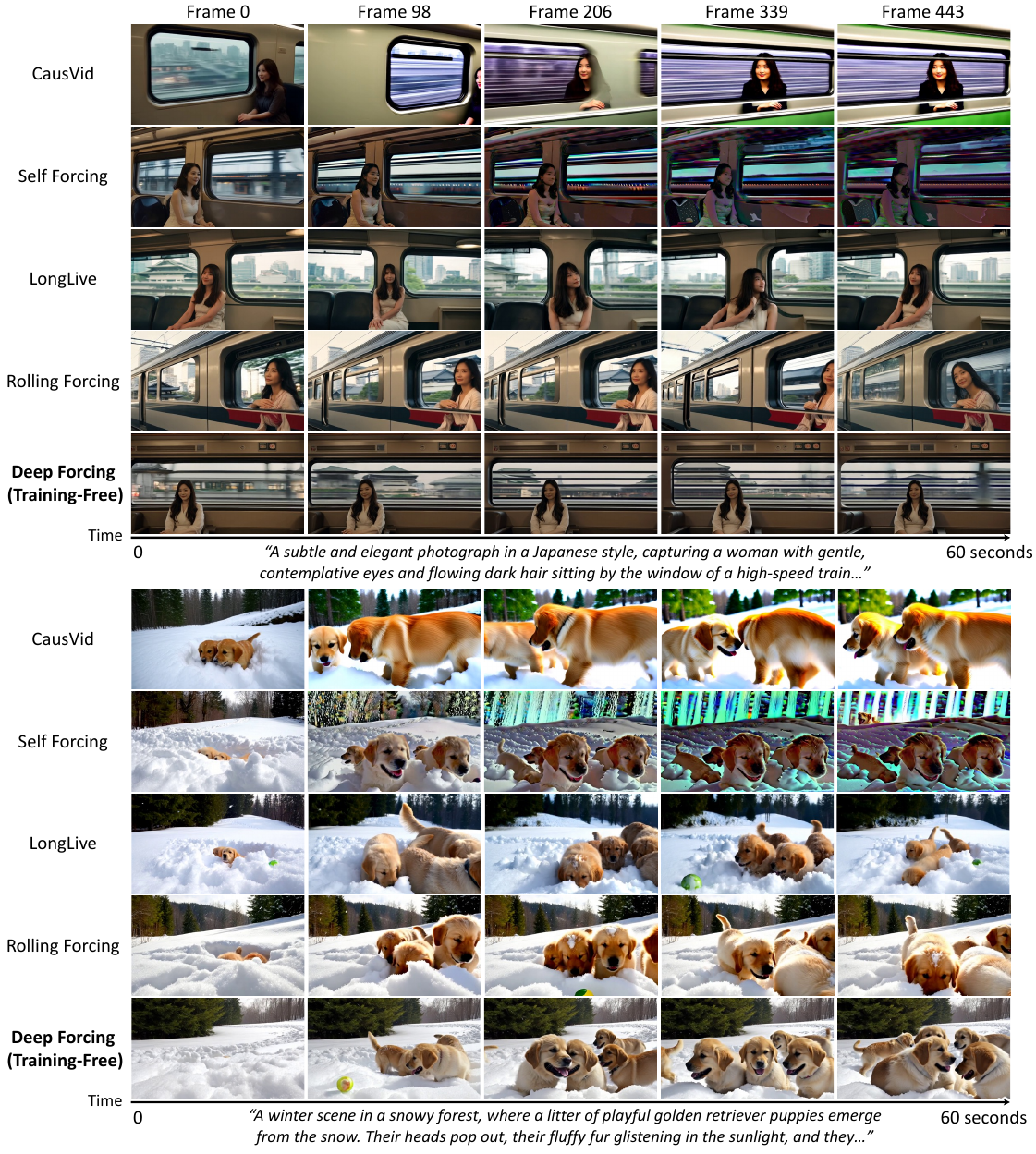}
  \caption{\textbf{Qualitative results on 60-second videos.} Frame-by-frame comparison across different methods for two representative prompts. Deep Forcing (training-free) achieves temporal consistency and visual quality comparable to training-based baselines (CausVid~\cite{yin2025slow}, Self Forcing~\cite{huang2025self}, LongLive~\cite{yang2025longlive}, Rolling Forcing~\cite{liu2025rolling}) while generating more dynamic content with greater subject consistency.}\vspace{10pt}
  \label{fig:suppl_qual2}
\end{figure*}

Additional qualitative results of our method are presented in Figure \ref{fig:suppl_qual1}, and Figure \ref{fig:suppl_qual2}. These examples clearly show that our \textbf{training-free} Deep Sink and Participative Compression framework produces results on par with training-based methods. 

\section{User Study Protocol}
\label{sec:userstudy}
To perform human evaluation, we conducted a user study based on the Two-Alternative Forced Choice (2AFC) protocol~\cite{blattmann2023stable, chefer2025videojam}. 
For each question, participants were presented with two videos generated from the same prompt and instructed to choose which video they preferred according to four evaluation criteria: (1) Color Consistency - Which video maintains more consistent color and exposure throughout, without sudden shifts in brightness, saturation, or color tone? (2) Dynamic Motion - Which video exhibits more dynamic and varied motion, including both subject movement and camera movement? (3) Subject Consistency - Which video maintains better visual consistency of the main subject throughout its duration? (Consider comparing the beginning and end of each video.) (4) Overall Quality - Overall, which video appears more realistic, natural, and of higher quality?

\begin{figure*}[t]
  \centering
\includegraphics[width=0.8\textwidth,clip]{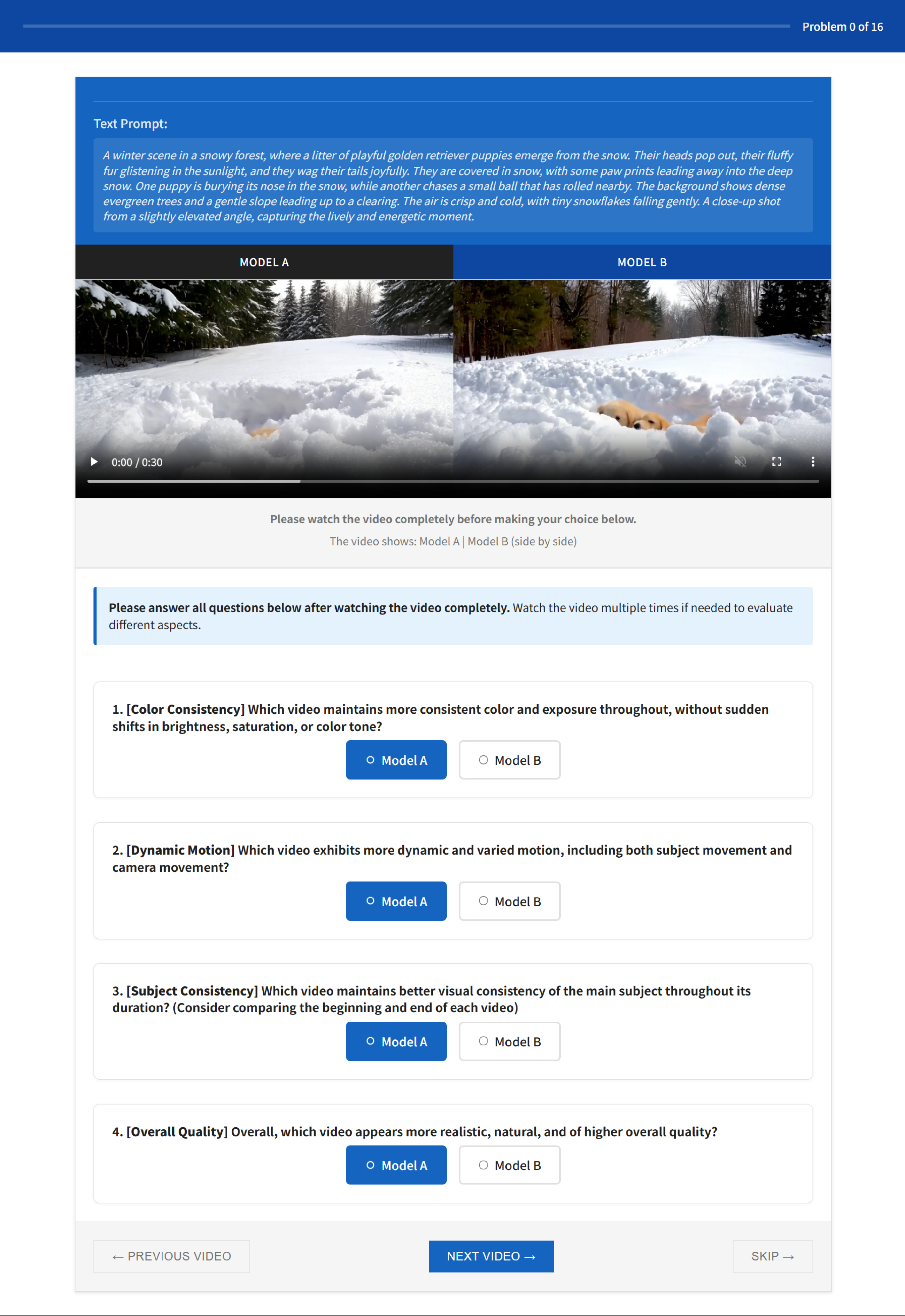}
  \caption{\textbf{Example of the user interface for the user study.} }\vspace{-10pt}
  \label{fig:user_study}
\end{figure*}

Each participant evaluated 16 video pairs comparing Deep Forcing against each of the four baselines (CausVid~\cite{yin2025slow}, Self Forcing~\cite{huang2025self}, LongLive~\cite{yang2025longlive}, and Rolling Forcing~\cite{liu2025rolling}). 
For each baseline, participants were shown 4 pairwise comparisons using different prompts, with all 16 prompts being non-overlapping within each participant's session. 
These prompts were randomly sampled from a pool of 20 total prompts. With 24 total participants, this yielded 384 total video comparisons (24 participants $\times$ 16 pairs), the results of which are shown in Table~\ref{tab:user_study}. 
The presentation order of videos was randomized, and participants were not informed which model generated each video. 
This design ensured balanced evaluation across all baseline models while avoiding prompt repetition within individual sessions. 
The user interface is shown in Figure~\ref{fig:user_study}.

\section{Additional Attention Visualization}
\label{sec:additional_attn_vis}

\begin{figure*}[t]
  \centering
\includegraphics[width=1.0\textwidth,clip]{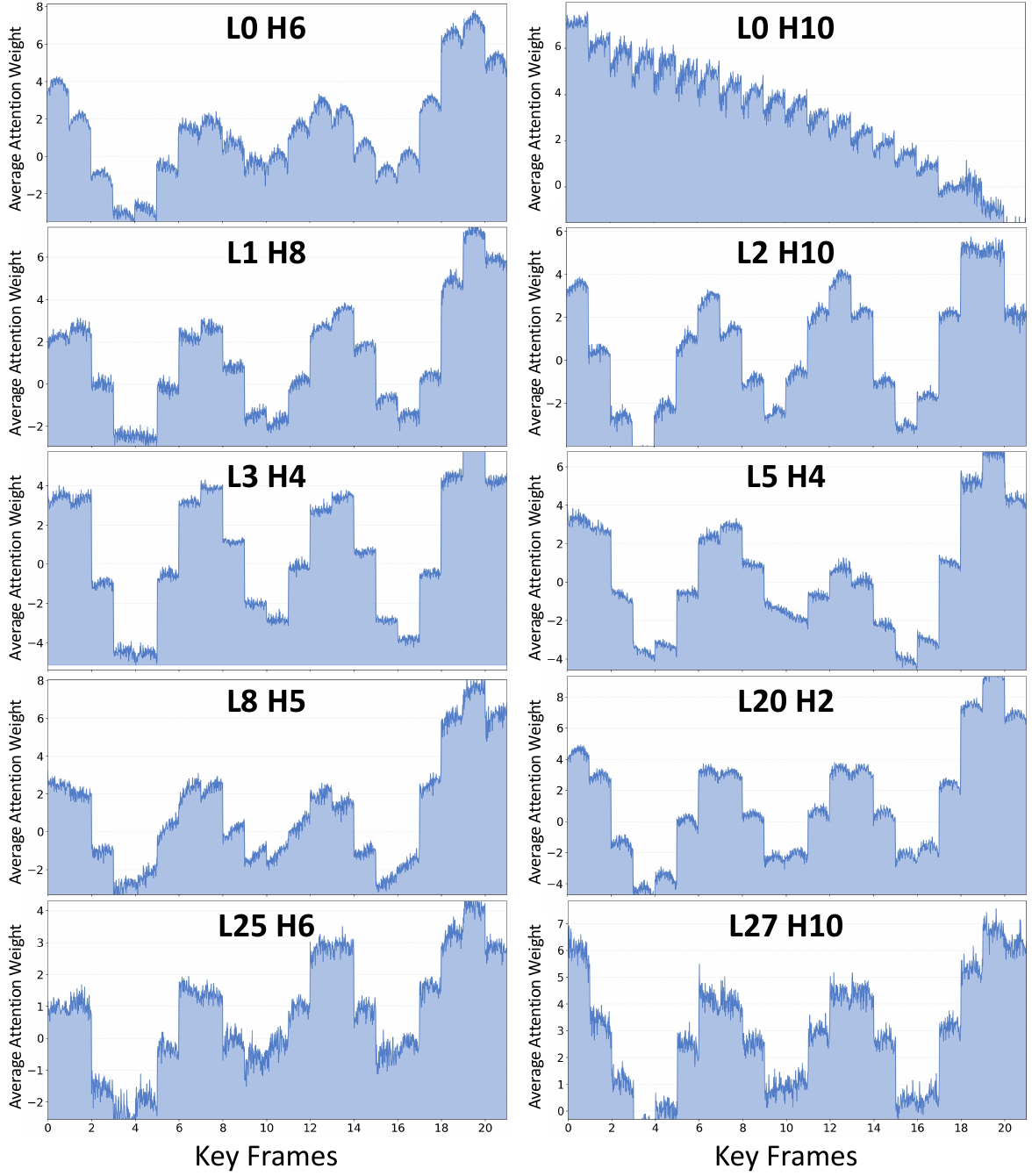}
  \caption{\textbf{Attention weight distribution across earlier frames.} Query-averaged attention showing how the last chunk (frames 19-21) attends to earlier KV cache entries (frames 0-18). Each frame consists of 1,560 tokens (spatially arranged latent patches). We visualize multiple attention heads from different layers, demonstrating that substantial attention to intermediate tokens is consistent across layers and heads.}\vspace{-10pt}
  \label{fig:additional_attention_visualization}
\end{figure*}

We provide additional attention head visualizations (Fig.\ref{fig:additional_attention_visualization}) beyond those shown in Fig.\ref{fig:attention_analysis} from Section~\ref{sec:deep-sink}. This deep attention pattern with substantial weight on both initial and intermediate tokens, emerges consistently and pervasively across layers and heads, rather than being only one or two specific heads, supporting the hypothesis that deep sinks are fundamental to Self Forcing~\cite{huang2025self}.

\end{document}